\documentclass[letterpaper, 10 pt, conference]{ieeeconf} 
\IEEEoverridecommandlockouts                              

\let\labelindent\relax 

\usepackage{cite}
\usepackage{amsmath,amssymb,amsfonts}
\usepackage{algorithmic}
\usepackage{graphicx}
\usepackage{subcaption} 
\usepackage{microtype}
\usepackage{textcomp}
\usepackage{xcolor}
\usepackage{float}
\usepackage{enumitem}

\usepackage{algorithm}

\begin{document}
\title{Multi-robot Multi-source Localization in Complex Flows with Physics-Preserving Environment Models} 


\author{Benjamin Shaffer$^{1\dagger}$, Victoria Edwards$^{2}$, Brooks Kinch$^{1}$, Nathaniel Trask$^{1}$, and M. Ani Hsieh$^{1}$
\thanks{*B. Shaffer's work is supported by the National Science Foundation Graduate Research Fellowship under Grant No. DGE-2236662. N. Trask's work is supported by SEA-CROGS (Scalable, Efficient and Accelerated Causal Reasoning Operators, Graphs and Spikes for Earth and Embedded Systems), a Mathematical Multifaceted Integrated Capability Center (MMICCs) funded by the Department of Energy Office of Science. M. A. Hsieh's work is supported by ONR Award \#N000142512171. The authors thank Quercus Hernandez for assistance designing Figure \ref{fig:method_overview}. }
\thanks{$^{1}$ Department of Mechanical Engineering and Applied Mechanics,
       University of Pennsylvania,
       Philadelphia, PA
       }%
\thanks{$^{2}$ Interdisciplinary Computing,
       College of the Atlantic,
       Bar Harbor, ME}%
\thanks{$\dagger$ Corresponding author, B. Shaffer:  \tt{\small{ben31@seas.upenn.edu}}}
}


\maketitle

\begin{abstract}

Source localization in a complex flow  poses a significant challenge for multi-robot teams tasked with localizing the source of chemical leaks or tracking the dispersion of an oil spill. The flow dynamics can be time-varying and chaotic, resulting in sporadic and intermittent sensor readings, and complex environmental geometries further complicate a team's ability to model and predict the dispersion.  To accurately account for the physical processes that drive the dispersion dynamics, robots must have access to computationally intensive numerical models, which can be difficult when onboard computation is limited. We present a distributed mobile sensing framework for source localization in which each robot carries a machine-learned, finite element model of its environment to guide information-based sampling. The models are used to evaluate an approximate mutual information criterion to drive an infotaxis control strategy, which selects sensing regions that are expected to maximize informativeness for the source localization objective. Our approach achieves faster error reduction compared to baseline sensing strategies and results in more accurate source localization compared to baseline machine learning approaches.

\end{abstract}

\section{Introduction}
\begin{figure}[h]
    \centering
    \includegraphics[width=0.5\textwidth]{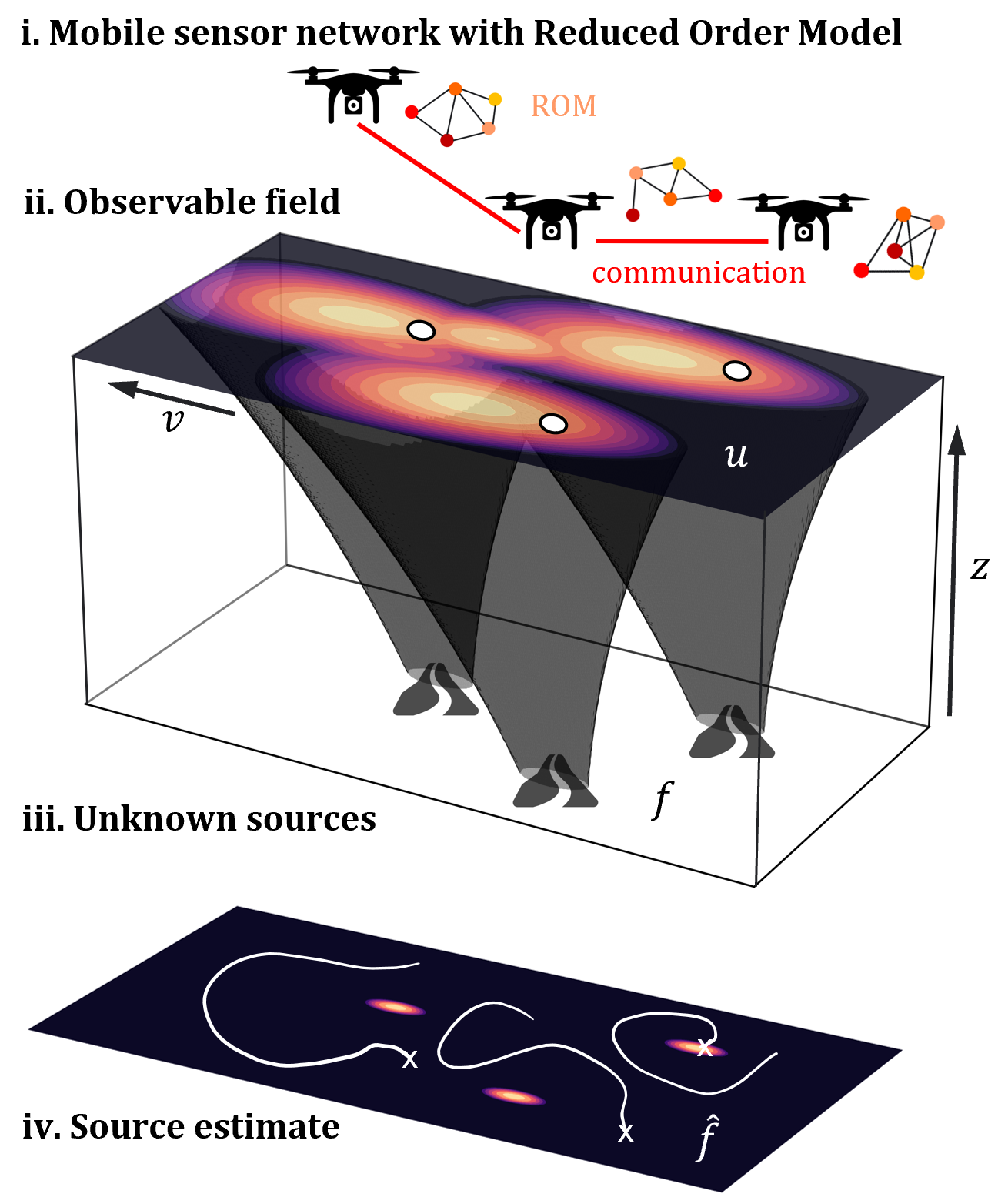}
    \caption{We consider a networked team of robots (i.) equipped with sensors measuring an observable field (ii.) resulting from the transport of an unobservable source (iii.). We equip each robot with a physics based reduced order model (i.) to perform source inference which are suitable for computing an estimate of the informativeness of potential sensing locations. The robots iteratively update their estimate and move to areas of higher sensing value to give a well resolved source estimate (iv.).}
    \label{fig:motivation}
\end{figure}

A core challenge in environmental monitoring is \emph{source localization}:
recovering the location of an unobservable source from sparse, indirect measurements \cite{ daw2022source, hon2010inverse}.
A robot team equipped with sensors is a promising solution because the team can adaptively select informative sampling locations and are robust for operations in hazardous conditions \cite{sung2023decision}. 
However, the robot team is often observing a field (e.g. concentration or temperature) resulting from the transport of a source under some physical process \cite{moghaddam2021inverse}.
For example, in hydrothermal vent identification, autonomous surface vehicles (ASVs) can directly detect the presence of ``tracer'' components on the ocean surface, but not the presence of an undersea vent\cite{yoerger2007autonomous}, {\it e.g.}, Fig. \ref{fig:motivation}. 
In this work, we address the problem of source localization through indirect measurements of a source field in the presence of a complex flow. 

The source localization problem is common in nature \cite{reddy2022olfactory}; the behavior of source-seeking moths inspired the infotaxis algorithm \cite{vergassola2007infotaxis}, where sensing locations are selected by greedily maximizing the expected informativeness for the source location, under a known prior model.
The method has been extended to robotics \cite{martin2010effectiveness}, in a distributed setting \cite{hajieghrary2017information}, for multiple sources \cite{chen2025dlw}, and in continuous time by treating the information as a potential field \cite{barbieri2011trajectories}. 
The key limitation of these approaches is the assumption of \textit{a prior} environmental models \cite{rodriguez2014limits}, and explicit consideration of a single source.
Solving complex fluid dynamics requires substantial computational resources and is often not feasible on resource-constrained robotic platforms.

To address this challenge, we utilize a structure-preserving machine learning framework \cite{kinch2025structurepreservingdigitaltwinsconditional, shaffer2025physicsinformedsensorcoveragestructure}, which serves as an adaptable, real-time environmental model, while maintaining key properties of the source transport dynamics (conservation, source to field coupling) \cite{kinch2025structurepreservingdigitaltwinsconditional}.
Structure preserving methods in machine learning are techniques which exactly preserve physical properties, for instance symmetries, invariances, or conservations by construction.
The Conditional Neural Whitney Form (CNWF) method we use combines the flexibility of transformer-based operator learning and the numerical properties of traditional numerical methods and guarantees global conservation. The conservation framing captures the relation between the observable concentration field and the source location through a learnable flux.
We use this model to derive an approximate mutual information metric, which can be computed locally onboard robots. This information metric captures the expected reduction in uncertainty of the source locations due to a measurement at a given location. To control the robot team, we employ an infotaxis-style information-gradient ascent controller \cite{barbieri2011trajectories}. Our approach builds on \cite{hajieghrary2017information} to enable a distributed implementation with inter-robot communication.
We model the source as a density which naturally allows for extension to an arbitrary number of local sources, or more complex distributed source.

We demonstrate that our approach produces more interpretable and effective sensing strategies than baseline sampling approaches in simulated settings.
More broadly, we view the integration of principled reduced order modeling and robot control as an example of onboard physical reasoning and highlight the future potential of well-designed surrogates for robot applications.
To that end, our contributions are:
\begin{itemize}[leftmargin=*]
    \item A structure preserving reduced order environment model for source localization; 
    \item An infotaxis-style information-based acquisition framework built around the learned environment model; and
    \item A distributed implementation using onboard computation.
\end{itemize}

\section{Related Work}
A common simplifying assumption in the source localization problem is that the source location coincides with the maximal value of the observable concentration field.
This results in gradient-based \cite{ogren2004cooperative, Talwar24}, gradient approximation \cite{atanasov2012stochastic}, and non-gradient based-methods for robot allocation \cite{al2021distributed, zhangQin2023distributed}. 
Factor graphs have also been used as the underlying environment model \cite{denniston2023fast}.
These approaches do not typically: address the effect of complex flows, provide direct estimates of the source term, and may not converge to the source, for instance if the concentration field is advected downwind, as in the hydrothermal vent example.

There are several recent approaches that incorporate explicit effects of advection-diffusion on the source signal.
For example, robots teams perform plume tracking using infotaxis-style solutions with different formulations \cite{hajieghrary2017information, wang2019dynamic, singh2023emergent}.
Various forms of Kalman filtering are combined with formation control to model spatiotemporal fields
\cite{zhang23, zhang2023distributed, mayberry2025soft} or perform level curve tracking \cite{guruswamy20}.
Proper Orthogonal Decomposition (POD) has been used to model the field and information theoretic planning for robot team sampling \cite{khodayi2019model}.
These works all make strict model assumptions for the underlying spatiotemporal process.
Our method is flexible to general nonlinear transport and learned from data while preserving the necessary source structure.

Machine learning presents general methods to improve environmental monitoring \cite{khodayi2018physics}.
Methods such as Physics Informed Neural Networks (PINNs)\cite{raissi2019physics} or operator learning \cite{lu2021learning, li2020fourier} cover a spectrum from incorporating explicit physics to purely data driven approaches and have been implemented for various inverse problems including source estimation\cite{de2022deep, mishra2022estimates, kamyab2022deep, vesselinov2018contaminant, kontos2022machine, reiter2017machine}.
However, these methods capture physical structure weakly (by penalty) if at all, and the coupling between the sampled field and the source term must be implicitly deduced from training \cite{karniadakis2021physics, wang2023physics}.

\emph{Structure preserving machine learning} has emerged as methods which incorporate physical or geometric structure directly, by construction. Broadly, these techniques can be understood as strictly enforcing a more fundamental level of physical properties (as opposed to particular equations as in PINNs), and have been shown to mitigate weaknesses of purely data driven approaches, particularly with data scarcity.
We build on recent methods designed to offer conservation by design, interpretable structure, and principled integration with mesh-based physical domains \cite{trask2022enforcing, actor2024data, jiang2024structure}.

\section{Problem Statement}

We consider a team of $N$ mobile robots operating in a bounded workspace $\Omega \subset \mathbb{R}^d$ with kinematics given by
\[
    \dot{x}_i = c_i, \qquad x_i \in \Omega,
\]
where $i = 1, \ldots, N$ denotes the $i^{th}$ robot and $c_i$ is the control input. Each robot carries a sensor measuring a scalar field $u:\Omega \to \mathbb{R}$ with additive noise:
\[
    d_i = u(x_i) + \varepsilon_i, \qquad \varepsilon_i \sim
    \mathcal{N}(0,\sigma^2).
\]

The field $u$ results from the distortion and displacement of an unknown source distribution $f:\Omega \to \mathbb{R}$ via a background flow. This process is defined as the forward operator $u = G(f)$. The set of measurements at a given time is $\mathcal{D} = \{(x_i,d_i)\}_{i=1}^{N}$.
The source estimation task is to identify $f$ from $\mathcal{D}$ while actively choosing robot motions to collect informative new measurements. This is also known as the \emph{inverse source identification} problem, and the operator $f=S(\mathcal{D})$ is known as the inverse operator.

We aim to design a control policy
\[
\dot{x}_i = \mathcal{P}(x_i,\mathcal{D})
\]
that drives the robots to collect data that enables accurate recovery of the unknown source field $f$ quickly and efficiently. Doing so requires knowledge of both the forward and inverse process, where $S$ is used to estimate the source from current or potential measurements, and  $G$ is needed to estimate the measurement along a potential robot trajectory for path planning.

\section{Methods}

\begin{figure*}[h]
    \centering
    \includegraphics[width=1.0\textwidth]{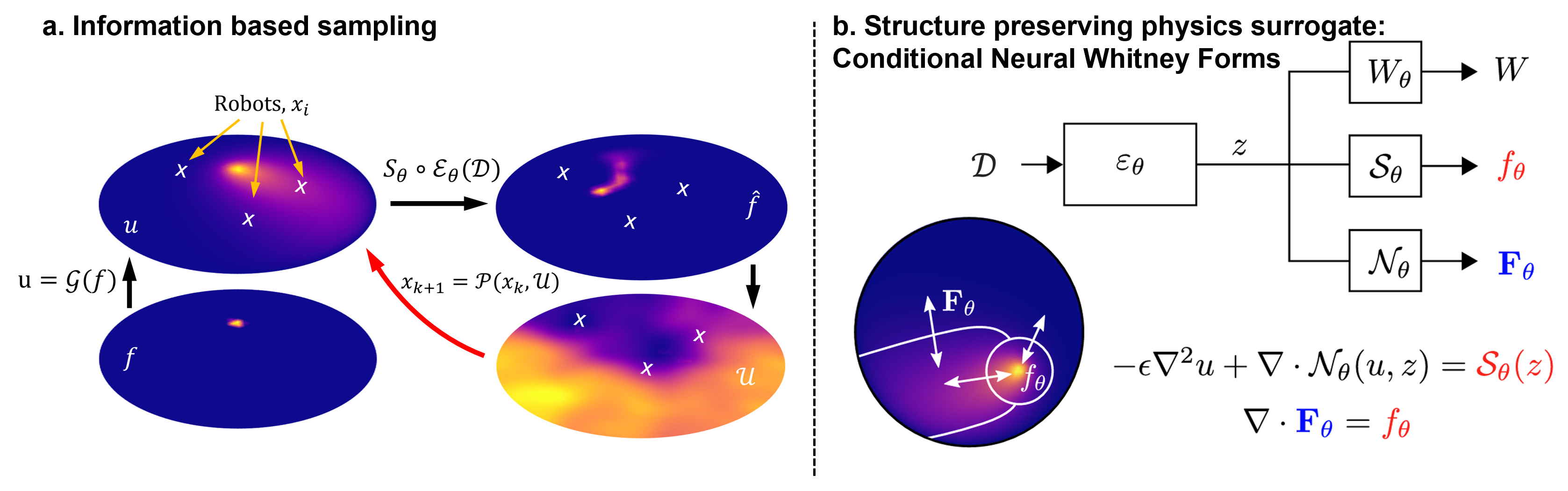}
    \caption{
    Method overview. We aim to infer the location of unobservable sources via targeted observations of the resulting scalar field, with a physics-based surrogate model driving the information based robot control strategy (shown in a.). The surrogate model is built on Conditional Neural Whitney Forms to provide a physically interpretable source model, discrete conservation by construction, and a geometric basis for handling complex geometries, in b.}
    \label{fig:method_overview}
\end{figure*}

We adopt an infotaxis-based approach to define the robot control policy. 
At each iteration, we score a candidate measurement location $x$ by its mutual information with the source field:
\[
    \mathcal{U}(x) := I\!\big[f;\,u(x) \mid \mathcal{D}\big].
\]
Finding paths for all robots to maximize the total information of future samples is a combinatorially hard collaboration problem.
As a computable proxy, we consider information gradient ascent infotaxis \cite{barbieri2011trajectories, julian2014mutual} as a method for directing robots to locally informative regions.
We, therefore, define the control policy as a gradient ascent over the information density
\[
    \mathcal{P}(x,\mathcal{D}) = \frac{1}{\gamma}\nabla\mathcal{U}(x,\mathcal{D})
\]
where $\gamma$ is a friction coefficient, defined as \(\gamma = 1/\lVert\nabla\mathcal{U}\rVert\).

Computing $\mathcal{U}(x)$ requires two key model components: (i) a forward operator $G$ to propagate candidate sources $f$ to predicted measurements $u$, and (ii) an inverse operator $S$ to infer a posterior distribution over $f$ from data $\mathcal{D}$. As exact PDE-based implementations of $G$ and $S$ are computationally prohibitive for online planning, we learn reduced order models $\hat{u}=G_{\theta}(f)$ and $\hat{f}=S_{\theta}(\mathcal{D})$ from data to enable real-time evaluation of $\mathcal{U}(x)$ and closed-loop control. This process is shown in Figure~\ref{fig:method_overview} (a).

\subsection{Bayesian Active Learning by Disagreement}
\label{sec:method_mi_comp}
In order to evaluate the mutual information objective, we express it as the
expected posterior entropy, and
rearrange as \cite{houlsby2011bayesian}
\begin{equation}
    \label{eq:BALD}
    \mathcal{U}(x) = \mathbb{H}[u \mid x,\mathcal{D}] - \mathbb{E}_{f\sim p(f \mid \mathcal{D})} \left[ \mathbb{H}[u\mid x,f ] \right], 
\end{equation}
which can be interpreted as seeking the $x$ for which the source prediction under the posterior leads to the most disagreement through the forward model and is known as Bayesian active learning by disagreement (BALD). 

To compute the mutual information objective, we maintain a posterior over the sources $p(f\mid \mathcal{D})=S_\theta(\mathcal{D})$ and conditional forward $p(u\mid f)=G_\theta(f)$. The composition of the inverse model and the forward operator defines the induced predictive posterior $p(u\mid \mathcal{D})=G_\theta \circ S_\theta(\mathcal{D})$.
We utilize dropout on the neural network models to obtain an approximately Bayesian model. While this is an imperfect method \cite{liu2021peril}, it has significant practical value in that it has theoretical backing \cite{gal2016dropout} and is very cheap to implement for inference in a batched context. In this sense, we prioritize compatibility with robot deployment and representation flexibility over more limited models that maintain a fully Bayesian posterior (Gaussian process), or ensemble methods which are expensive to evaluate in real time.

\begin{figure}[htbp]
    \centering
    \includegraphics[width=0.45\textwidth]{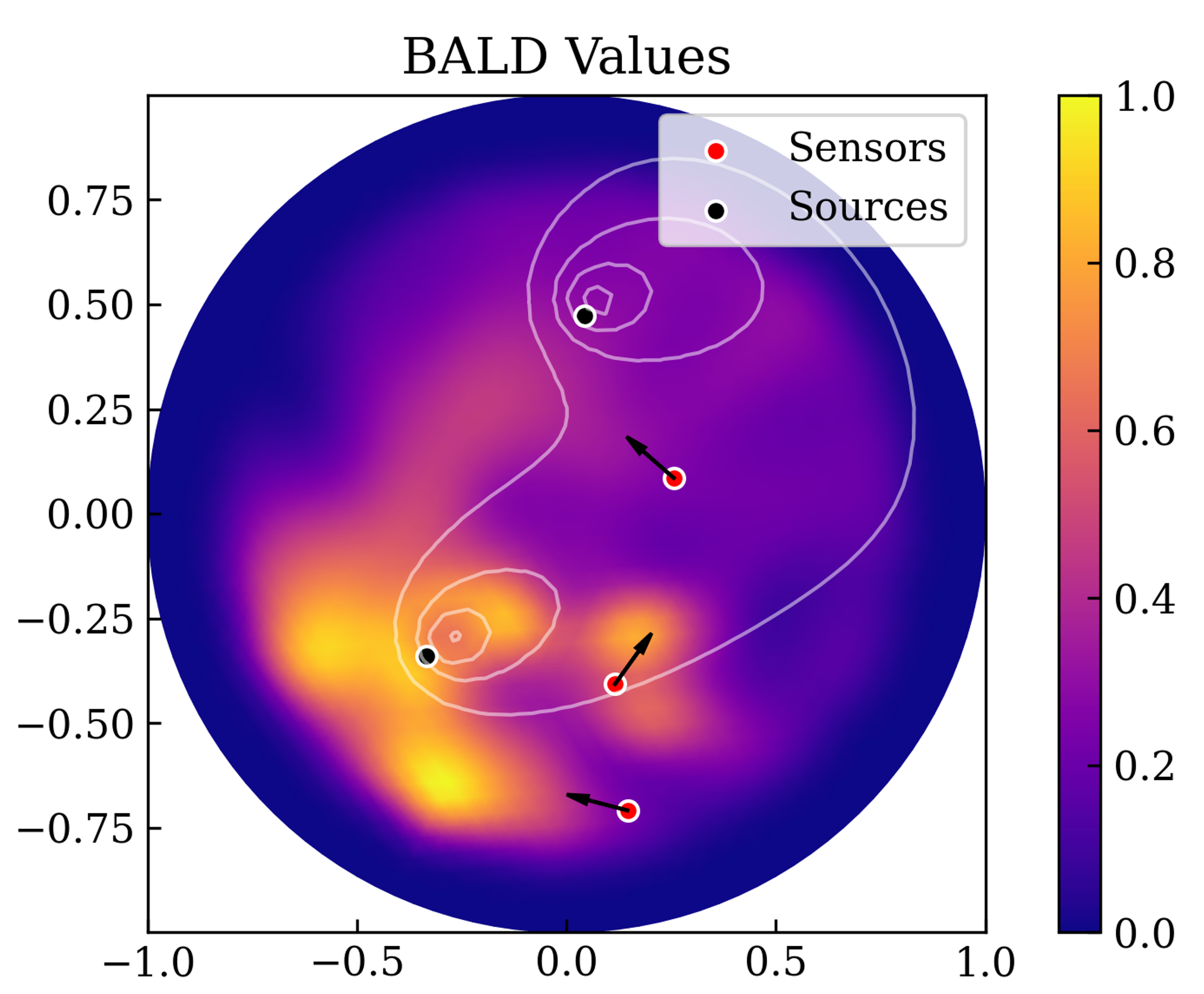}
    \caption{We demonstrate the evaluation of $\mathcal{U}$ for a representative configuration. Arrows indicate the direction of the information gradient at each robot location (red) which is used in the local and distributed implementation. The information based acquisition function identifies upstream and unobserved locations as the most informative, matching expectations. The observable scalar field is shown in a contour.}
    \label{fig:bald_over_domain}
\end{figure}

We approximate the BALD acquisition utility for a candidate location $x \in \Omega$ using Monte Carlo sampling from the learned probabilistic inverse and forward models. An example evaluation of $\mathcal{U}$ is shown in Figure~\ref{fig:bald_over_domain}.


\subsection{Structure Preserving Reduced Order Model}
\label{sec:method_CNWF}
Using the recently developed Conditional Neural Whitney Forms (CNWF) method \cite{kinch2025structurepreservingdigitaltwinsconditional} we define the forward operator $G_\theta$ and inverse operator $S_\theta$ needed for robot motion planning and source estimation.
This approach is built on finite element methods to provide an adaptable PDE model with exact conservation, conditioned on real-time sensor measurements \cite{shaffer2025physicsinformedsensorcoveragestructure}. 
An overview of this architecture is shown in Figure~\ref{fig:method_overview} (b). 

There are four learnable components.
First, an encoder, 
\begin{align*}
    \label{eq:encoder}
    z = \mathcal{E}_\theta(\mathcal{D}),
\end{align*} 
which maps sensor readings to a latent representation. 
Then three networks conditioned on $z$,
a \emph{source prediction} model, $\hat{f}=S_\theta(x)$,  a \emph{flux} model, $F=\mathcal{F}_\theta(z,u)$, which describes the differential operator, and an \emph{adaptable reduced basis}, $W=W_\theta(z)$, which all quantities are represented on.
Together, these describe a discrete conservation of the form
\begin{equation}
    \label{eq:CNWF-PDE}
    \mathcal{H}(u,z):=\nabla \cdot \mathcal{F}_\theta(u, z) - S_\theta(z) = 0.
\end{equation}
where the encoded observations, $z$, act to calibrate; the learned differential operator, adaptive basis, and source term in real-time.
This results in a PDE model which best matches the observed data. 
Through a careful discretization using finite element exterior calculus (FEEC)-- a framework developed for electromagnetism-- we can ensure this conservation is satisfied exactly \cite{arnold2018finite}. 
The benefits of this framework for our application are
\begin{itemize}[leftmargin=*]
    \item The reduced basis allows each robot to evaluate their \textbf{lightweight finite element model}.
    \item The learned flux \textbf{couples scalar field observations to the source distribution} during training.
    \item The PDE operator defines a forward map, $G_\theta$, for computing information objective.
    \item The learned source term is \textbf{physically interpretable}.
    \item Offline training allows models to be preloaded onto robots.
    \item Flexible online adaptation via conditioning allows for \textbf{distributed implementation}.
\end{itemize}

For a given encoded observation input $z$, the models are evaluated and \eqref{eq:CNWF-PDE} is solved using a standard nonlinear FEM Newton solver to yield the solution $\hat{u}$ \cite{kinch2025structurepreservingdigitaltwinsconditional}. We identify this forward solve as the parameterized forward operator, 
\[
    G(f):=\hat{u}\quad \text{such that} \quad \mathcal{F}(\hat{u},z)=f
\]
(a solve of the learned PDE) and the source model $f_\theta$ as $S_\theta$.

\subsubsection{Training procedure}
We establish an offline-training, online-calibration procedure. Prior to use we assume each robot is equipped with a fully trained model (consisting of each trainable component), which is trained offline on representative data on the same domain. Online we perform inference by calibrating the learned PDE to sensor measurements and evaluate this model to generate the mutual information metric.

We define the offline, supervised training process. 
Given a training set
\[
\mathcal{D}_{\mathrm{train}}
  = \big\{(\hat{\mathcal{D}}^{(j)},\,f^{(j)},\,u^{(j)})\big\}_{j=1}^{N},
\]
where $\hat{\mathcal{D}}^{(j)}$ are synthetic observations derived from $u^{(j)}$, 
we jointly learn $G_{\theta}:f \mapsto \hat{u}$ and $S_{\theta}:\mathcal{D}\mapsto \hat{f}$ 
by minimizing a reconstruction and consistency loss, subject to the learned constraint:
\begin{align}\label{eq:opt}
    \min_{\theta} \; &\| \hat{u} - u_{\text{true}} \|^2 + W_{2,\epsilon}^2(\hat{f}, f_{\mathrm{true}})
    \\
    &\text{subject to} \quad \mathcal{H}(\hat{u}, \mathcal{D}) = 0,
\end{align}
where $W_{2,\epsilon}^2$ is the Wasserstein transport metric. This is a PDE constrained optimization where $\hat{u}$ is defined implicitly through solving the constraint, ensuring conservation is satisfied exactly, independent of data scarcity.

\subsubsection{Implementation details}
CNWFs are a mesh-based method and therefore we assume access to a discretization of the domain $\Omega^h$, under basis functions $\{\varphi_i\}_{i=1}^{n_f}$, where $n_f$ is the discrete problem dimension.
Following \cite{shaffer2025physicsinformedsensorcoveragestructure}, we use an encoding self-attention transformer model to provide a permutation invariant and rank adaptable representation of the sensing inputs, \(z = \mathcal{E}_\theta(\mathcal{D}).\)
We define the parameterized flux in \eqref{eq:CNWF-PDE} as
a combination of a discrete Laplacian and a learned flux, \(\mathcal{N}_\theta(u, z)\), and a parameterized source as \(S_\theta(z)\)
and obtain the learned PDE form:
\begin{equation}
    \label{eq:learned-pde}
    -\eta \Delta u + \nabla \cdot \mathcal{N}_\theta(u, z) = S_\theta(z) \quad \text{in } \Omega, \quad u|_{\partial \Omega} = g,
\end{equation}
where $\eta$ is a user chosen parameter to ensure stability.
We define coarse basis functions $\{ \psi_j \}_{j=1}^{n_c}$ as convex combinations of the fine-scale basis $\varphi_i(x)$:
\begin{equation}
    \psi_j(x) = \sum_{i=1}^{n_f} W_{ji}(z)\, \varphi_i(x),
\end{equation}
where $W(z) \in \mathbb{R}^{n_c \times n_f}$ is a learned, nonnegative matrix.
This exposes a fully connected graph structure, with the learned fluxes operating on the edges.
All quantities are projected into this reduced basis.
The system is shown to be well-posed under reasonable continuity and regularity assumptions \cite{kinch2025structurepreservingdigitaltwinsconditional}.

\subsection{CNWF-BALD}

Combining the CNWF model defined in Section \ref{sec:method_CNWF} and the mutual information computation in Section \ref{sec:method_mi_comp}, we define our robot control policy using information gradient ascent. Algorithm \ref{alg:cnwf-bald} describes the combination of these methods. 
The spatial gradient can be readily computed on the finite element basis used for the CNWF model. In Section \ref{sec:res_adaptive_placement} we demonstrate that this method results in good robot placement under global waypoint control and local gradient ascent policies.

\begin{algorithm}
  \caption{CNWF--BALD Active Sensing}
  \label{alg:cnwf-bald}
  \begin{algorithmic}[1]
    \REQUIRE Pretrained CNWF model $(\mathcal{E}_\theta,\,S_\theta,\,G_\theta)$
    \REQUIRE Initial robot locations $X^{(0)}$, initial measurement history $\mathcal{D}^{(0)}$, iteration count $K$, time step $\Delta t$
    \REQUIRE Control policy $\mathcal{P}(\cdot)$
    \ENSURE Final source estimate $\hat{f}_K$ and robot trajectories $\{X^{(k)}\}_{k=0}^{K}$
    \FOR{$k = 0$ \TO $K-1$}
      \STATE \textbf{Sensing:} $d_i^{(k)}$ at $x_i^{(k)}$, update $\mathcal{D}^{(k+1)}$
      \STATE \textbf{Encode:} $z^{(k+1)} \gets \mathcal{E}_\theta(\mathcal{D}^{(k+1)})$
      \STATE \textbf{Source estimate:} $\hat{f}_k \gets S_\theta(z^{(k+1)})$
      \STATE \textbf{Compute MI:} $\mathcal{U}(x)$, using \eqref{eq:BALD}
      \STATE $X^{(k+1)} \gets \Delta t \;\mathcal{P}(\mathcal{U},X^{(k)})$
    \ENDFOR

    \STATE $\hat{f}_K \gets S_\theta(\mathcal{E}_\theta(\mathcal{D}^{(K)}))$, final source estimate
    \RETURN $\hat{f}_K,\,\{X^{(k)}\}_{k=0}^{K}$
  \end{algorithmic}
\end{algorithm}

\subsubsection{Distributed implementation}

Centralized coordination is often infeasible in multi-robot settings due to limited bandwidth, latency, and single points of failure. We therefore use a fully distributed architecture in which each robot maintains its own local copy of the CNWF surrogate, performs local inference, and communicates information with neighbors. We perform one communication step per active sensing iteration, $k$.

Let $\mathcal{H}^{(k)}$ be the set of neighboring robots at iteration $k$. We define neighbors as all robots within a fixed communication radius $r_c$:
\begin{equation}
    \label{eq:neighbor_def}
    \mathcal{H}_i^{(k)}:=\{j\; | \; \lVert x_i-x_j\rVert_2<r_c, \; j=1,\dots,N  \}.
\end{equation}
Each robot maintains a local history of known observations consisting of both collected and communicated data,  $\mathcal{D}_i^{(k)}$.
During a communication step, each robot shares its current measurement $\{d_i^{(k)},x_i^{(k)}\}$, and its history of measurements $\mathcal{D}_i^{(k)}$, with all neighbors. 
Incoming communications are assimilated into each robot's history:
\begin{equation}
    \label{eg:comm_update}
    \mathcal{D}_i^{(k+1)} \gets \mathcal{D}_i^{(k)} \bigcup_{j\in\mathcal{H}_i^{(k)}} \mathcal{D}_j^{(k)} \cup \{d_j^{(k)},x_j^{(k)}\}.
\end{equation}
We assume single-hop, synchronous communication which means information can only propagate one edge per time step, preserving causality in the network.
Each robot then solves its own forward problem using the CNWF surrogate conditioned on $\mathcal{D}_i^{(k+1)}$ to obtain estimates $\hat u_i(x),\hat f_i(x)$ and an information field $\mathcal{U}_i(x)$. This process is described in Algorithm \ref{alg:distributed-cnwf-bald}

\begin{algorithm}
  \caption{Distributed CNWF--BALD with Local History Fusion}
  \label{alg:distributed-cnwf-bald}
  \begin{algorithmic}[1]
    \REQUIRE Pretrained CNWF model $(\mathcal{E}_\theta,\,S_\theta,\,G_\theta)$
    \REQUIRE Initial robot locations $X^{(0)}$, initial measurement histories $\mathcal{D}_i^{(0)}$, iteration count $K$, communication radius $r_c$, time step $\Delta t$
    \REQUIRE Local control policy $\mathcal{P}(\cdot)$
    \ENSURE Final source estimates $\hat{f}_K$ and robot trajectories $\{X^{(k)}\}_{k=0}^{K}$
    \FOR{$k=0$ \TO $K-1$}
        \FOR{$i=1$ \TO $N$}
          \STATE \textbf{Sensing:} $d_i^{(k)}$ at $x_i^{(k)}$
          \STATE \textbf{Establish Communication}: Update $\mathcal{H}_i^{(k)}$ by \eqref{eq:neighbor_def}
          \STATE \textbf{Communication:} Update $\mathcal{D}_i^{(k+1)}$ by \eqref{eg:comm_update}
          \STATE \textbf{Encode:} $z^{(k+1)} \gets \mathcal{E}_\theta(\mathcal{D}_i^{(k+1)})$
          \STATE \textbf{Source estimate:} $\hat{f}_{k,i} \gets S_\theta(z^{(k+1)})$
          \STATE \textbf{Compute MI:} $\mathcal{U}_{i}(x)$, using \eqref{eq:BALD}
          \STATE \textbf{Control Update:} $x_i^{(k+1)} \gets \Delta t \;\mathcal{P}(\mathcal{U}_{i}, x_i^{(k)})$
        \ENDFOR
    \ENDFOR
    \STATE $\hat{f}_{i,K} \gets S_\theta(E_\theta(\mathcal{D}_i^{(K)}))$, final local source estimates
    \RETURN $\hat{f}_{i,K},\,\{X_i^{(k)}\}_{k=0}^{K}, \; i=1,\dots N$
  \end{algorithmic}
\end{algorithm}

This protocol yields a monotone growth of each robot's information set over time, approaching the global dataset after sufficient communication steps while preserving strictly local, causal updates at each round. The approach is robust to intermittent communication failures and scales naturally to large robot teams \cite{dimakis2010gossip, bullo2009distributed}.
In Section \ref{sec:dist_impl}, we empirically verify that i.) measurements effectively propagate through the robot network over the adaptive sampling run, and ii.) the average local prediction with partial observation inputs converges to an idealized centralized prediction made with all measurements.

\subsection{Experimental setup}
For an  example system we consider two dimensional steady-state advection-diffusion. While this problem is a linear PDE, for an unknown velocity, the corresponding inverse problem is nonlinear. The PDE is: 
\begin{equation}
    \label{eq:advec-diff}
    - \frac{1}{\text{Pe}} \nabla^2 u + \mathbf{v} \cdot \nabla u = f
    \quad \text{in } \Omega,
    \qquad
    u|_{\partial \Omega} = g,
\end{equation}
where $\text{Pe} > 0$ is the Peclet number,
$\mathbf{v}: \Omega \to \mathbb{R}^d$
is a potentially unknown, divergence-free velocity field with $|\mathbf{v}|=1$, $g$ is a fixed boundary condition.
To generate data we use a prescribed velocity and diffusivity. The system is solved on the discretized domain used in the CNWF model, $\Omega^h$.

We consider two scenarios in this work, first a circular geometry with a constant velocity with a randomly selected orientation (randomly chosen for each trial), and second a complex geometry with a variable stokes-flow velocity profile based on two inlets and six possible outlets, any combination of which may be open for a given sample (chosen randomly). Both the discretization $\Omega^h$ and boundary conditions $g$ are provided to initialize the CNWF model.
We generate data by solving the advection diffusion equations on the problem domain using finite elements. For each solution, we consider $n \sim U[1,5]$ sources randomly distributed throughout the domain. We use a constant Peclet number of $1e1$.

\section{Results}

\subsection{Source Reconstruction}

We evaluate the ability of the CNWF reduced order model to reconstruct the source field \( f \) from sparse observations.
We compare our method against a transformer-based baseline trained to regress from sensor data to the source field without imposing any conservation structure. The baseline has an identical architecture to the CNWF encoder $\mathcal{E}_\theta$ and a prediction head, for fair comparison.

Table~\ref{tab:recon} reports the mean Wasserstein-2 distance between the predicted source field and ground truth over 100 random test instances. CNWF achieves significantly lower reconstruction error, indicating that embedding physical structure into the surrogate improves accuracy and generalization in data-sparse regimes.

\begin{table}[ht]
    \centering
    \begin{tabular}{c|c|c}
        Geometry & Model &  $W_2(\rho_\theta^0, \rho_{true})$ $\downarrow$ \\
        \hline
        \textbf{Circle} & CNWF (\textit{ours}) & \textbf{7.12e-3} \\
         & Transformer (\textit{baseline}) &  2.92e-2 
        \\
        \hline
        \textbf{Hallway with Rooms} & CNWF (\textit{ours}) & \textbf{2.48e-1}  \\
         & Transformer (\textit{baseline}) &  3.26e-1 
        
    \end{tabular}
    \caption{Source prediction accuracy comparison between CNWF model and baseline transformer}
    \label{tab:recon}
\end{table}

\subsection{Active sensing via mutual information}
\label{sec:res_adaptive_placement}
\begin{figure}[htbp]
    \centering
    \includegraphics[width=0.4\textwidth]{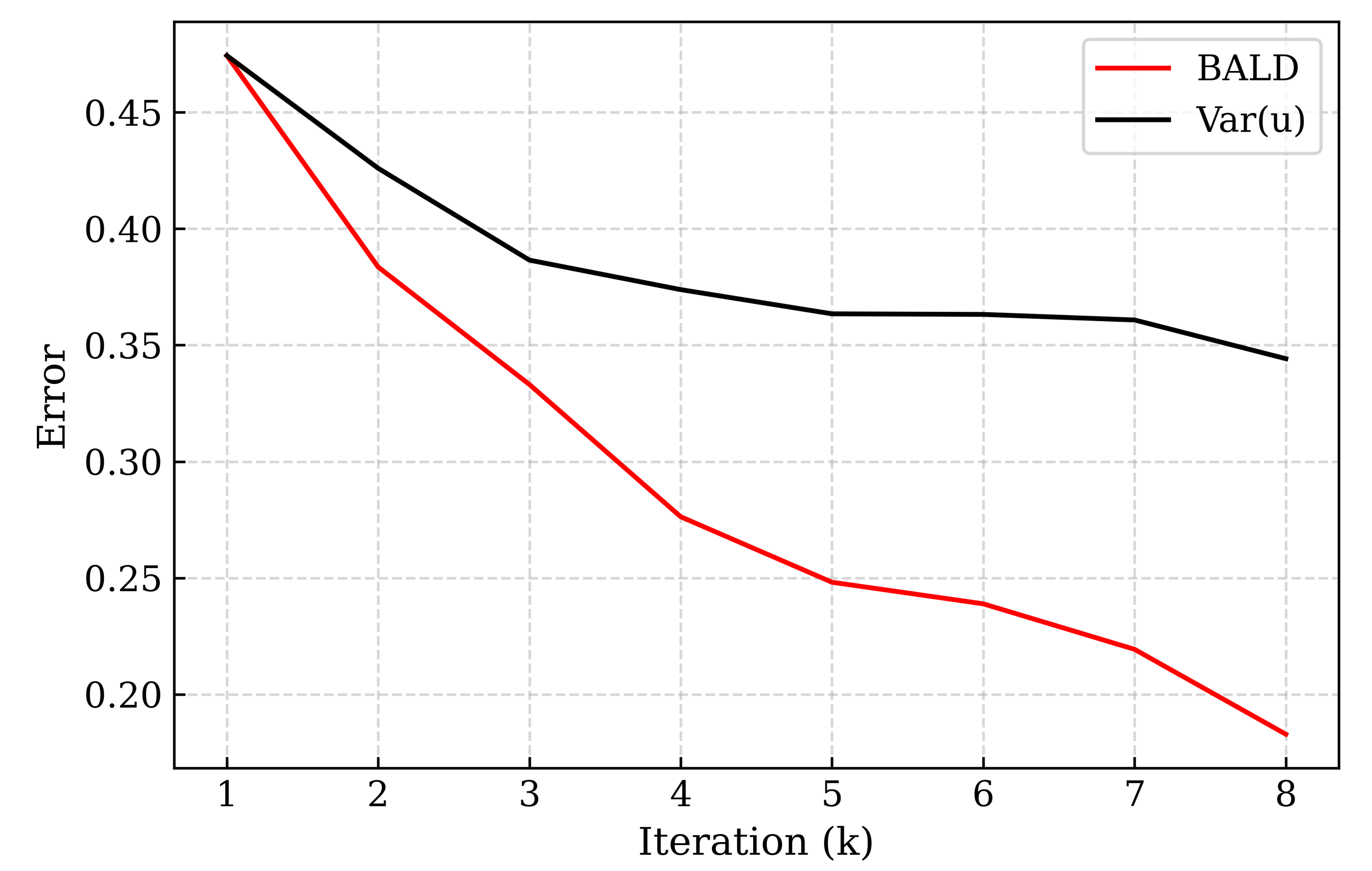}
    \caption{We compare the average source reconstruction error for 128 trials over $K$ iterations using global sensor placement strategies for the $\mathcal{U}$ and a naive variance based baseline $\textrm{Var}(u)$ on the circular domain. Our approach results in more informative sensor placement which in turn substantially reduces source error compared to the baseline.}
    \label{fig:BALD_var_comp}
\end{figure}

We compare our information-based robot control strategy against a naive variance maximizing baseline:
\[
    \mathcal{U}_{var}(x) = \sigma_u(x).
\]
This is a popular method due to its simplicity to implement, which treats uncertainty as a proxy for informativeness, and does not address the inverse source identification problem, i.e. this approach only requires a model for the scalar field $u$.
Most previous source seeking approaches (e.g. gradient or coverage-based) assume direct observation of the field of interest and do not provide direct comparison to our method which addresses the inverse source identification problem.
For a global acquisition strategy, we simply select the optimal sensing location as
$
    x^* = \text{argmax}_x\; \mathcal{U}(x).
$
We compare the information and variance based acquisition functions under the global waypoint robot control policy. Both methods use an identical CNWF model. We record the average source reconstruction error according to the Wasserstein metric over 8 iterations for 128 initial configurations in the circular domain in Figure~\ref{fig:BALD_var_comp}. Our method consistently reduces the reconstruction error more rapidly than the variance based baseline, demonstrating the value of information-driven placement. The variance based sampling methods requires 5 iterations to improve over a single iteration of the BALD method.

\subsubsection{Deployment in complex geometries}
\label{sec:complex_geom}
\begin{figure}[htbp]
    \centering
    \includegraphics[width=0.4\textwidth]{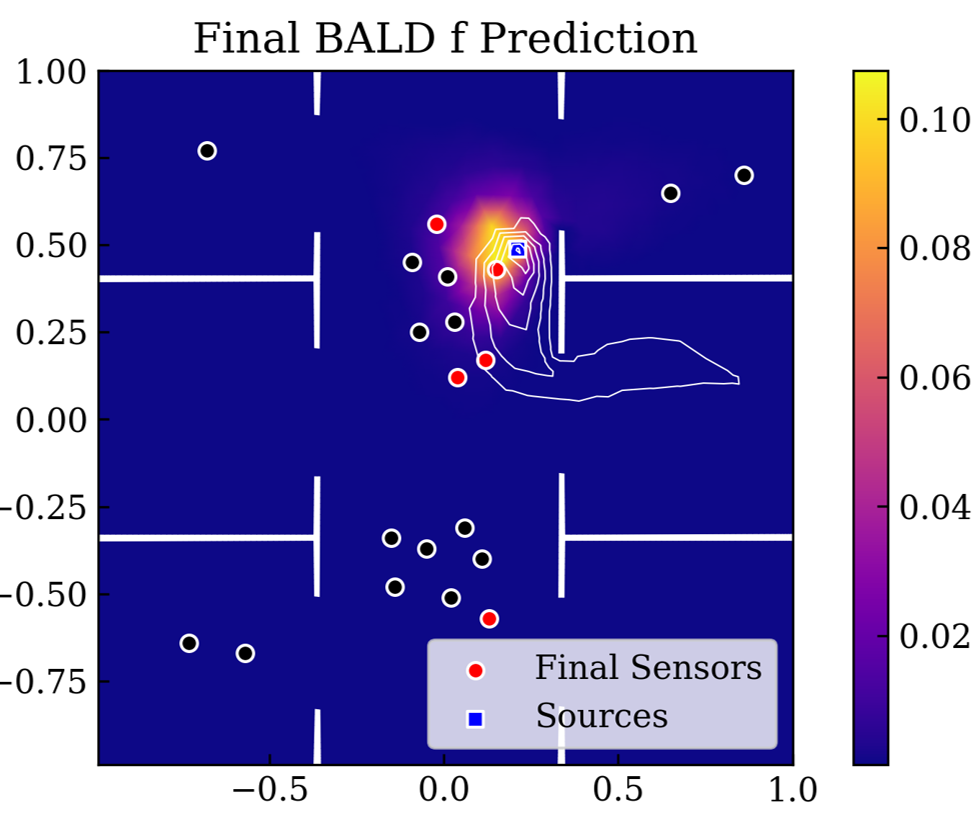}
    \caption{The global placement strategy on the hall geometry results in accurate localization of the source from an ill-posed initial configuration. The color indicates the source predictions and the contour indicates the observable scalar field.}
    \label{fig:hall_bald}
\end{figure}

We demonstrate the ability of the CNWF model to recover the source in a closed loop prediction context for the complex geometry with the CNWF-BALD global acquisition function in Figure~\ref{fig:hall_bald}. Interestingly, while most sensors are allocated near the source, some are also placed in the other intersection in the hallway, which qualitatively would seem to be an informative location due to the flow variability.

\subsection{Local Exploration via Information Gradient Ascent}
\begin{figure}[htbp]
    \centering
    \includegraphics[width=0.40\textwidth]{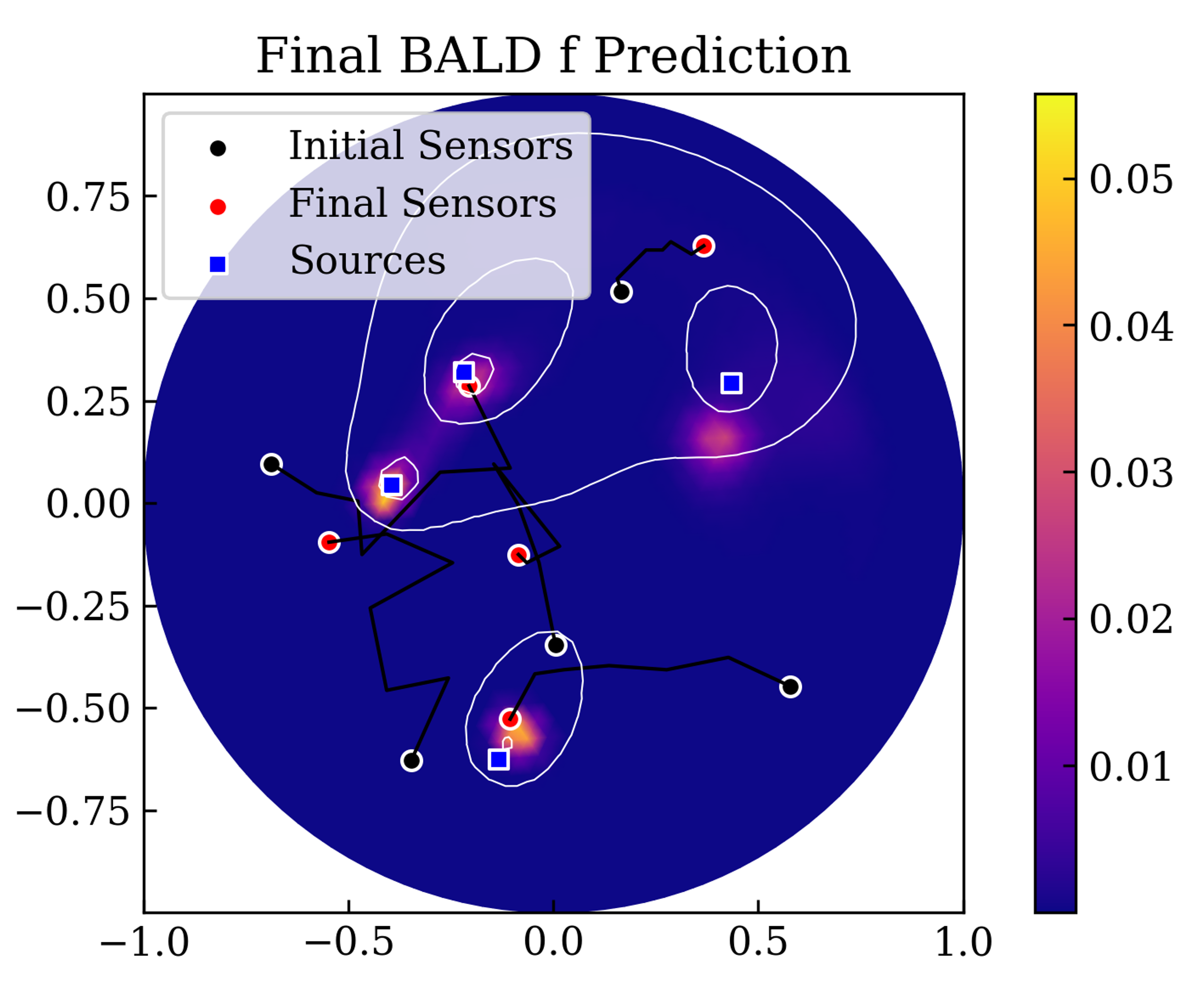}
    \caption{A sample gradient ascent demonstration for four unknown sources. The background color indicates the predicted source term, the contour indicates the scalar field, and the black lines shows the robot trajectories. In the final configuration (after 6 steps) the CNWF model correctly identifies the presence of four sources and accurately places three of them. The gradient ascent results in well distributed sensors and qualitatively good sensing locations near sources.}
    \label{fig:gradient_ascent_demo}
\end{figure}

We also evaluated the local, information-gradient ascent implementation of the infotaxis algorithm.
Figure~\ref{fig:gradient_ascent_demo} shows an example of this strategy for simultaneously localizing four sources.
In only three adaptive sensing iterations the source reconstruction error was reduced by 86\%.
Qualitatively, the information-based control policy results in a combination of both exploration and exploitation like behaviors, where some robots move directly to a nearby source and others move towards unsampled regions of the domain. The final configuration shows a spatially distributed robot configuration, even without the inclusion of an explicit repulsion in the control. Empirically, this result holds over all cases where the sensors are initially distributed.
The model successfully recovers all sources distributed over the domain with only 15 total sampling locations. While this local method does not guarantee globally optimal placement, it achieves competitive source reconstruction with reduced sampling overhead and realistic constraints. 

\subsection{Distributed Implementation}
\label{sec:dist_impl}

\begin{figure}[htbp]
    \centering
    \includegraphics[width=0.4\textwidth]{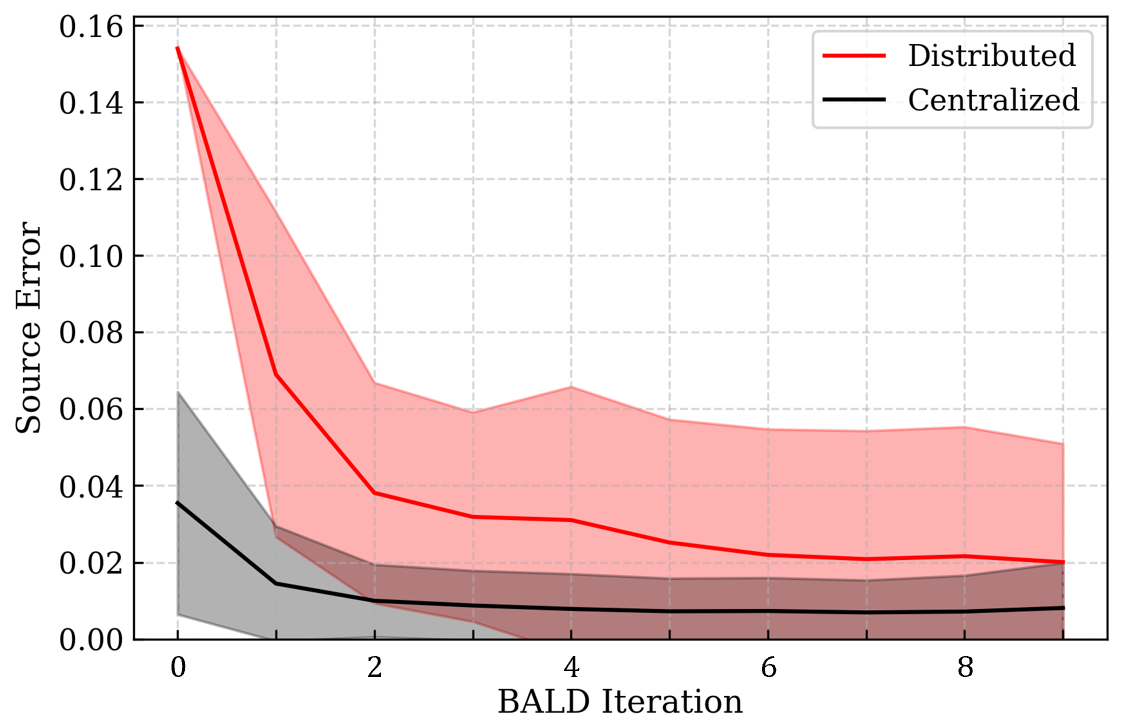}
    \caption{We compare the prediction error for each agent in the distributed implementation with an idealized centralized model. While initially the sparse communication results in degraded performance for the distributed implementation, error greatly improves as information propagates the sensor network.}
    \label{fig:dist_demo}
\end{figure}

\begin{figure}[htbp]
    \centering
    \includegraphics[width=0.4\textwidth]{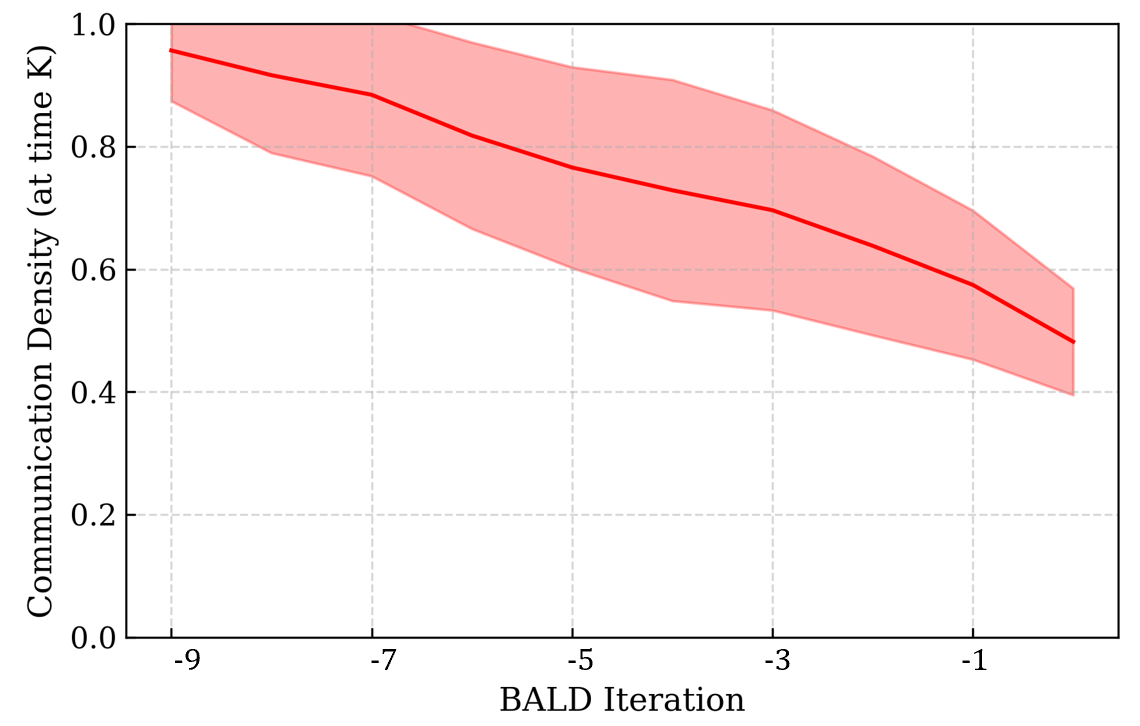}
    \caption{At the final iteration $K$, we evaluate the density of information from each previous active sampling step $-k$, that has reached each agent. While recent measrurements $k=0$ are only about $50\%$ dense, while nearly all observations made at time $k=-9$, have reached all robots at the current time. This shows the effectiveness of our simple communication protocal for collating observations.}
    \label{fig:dist_comm}
\end{figure}

Our experiments in Figure \ref{fig:dist_demo} demonstrate that the distributed CNWF-BALD algorithm achieves performance comparable to the centralized planner. With some exceptions, the collection and exchange of measurements results in increasingly consistent predictions. However, isolated robots may fail to reconnect, a known pitfall of these types of collaboration strategies.
We also observe that early measurements effectively propagate through the sensor network in Figure \ref{fig:dist_comm}.

\section{Conclusions}
We presented a framework for distributed multi-robot, multi-source localization in which each robot is equipped with a machine-learned reduced-order model of its environment. Leveraging the PDE-structured nature of these models, we derived an information-theoretic objective that drives informative path planning via gradient-based control. We further developed a communication and control strategy suitable for fully distributed operation. This approach unifies ideas from Bayesian optimization, PDE-constrained modeling, operator learning, and robotics, achieving improved source estimation and more informative sampling compared to baseline machine learning and planning methods. More broadly, equipping robots with onboard, structure-preserving finite element surrogates demonstrates the potential for robots to act as physically informed, model-driven agents in scientific exploration and sensing tasks.

\bibliographystyle{IEEEtran}
\bibliography{bib}

\begin{thebibliography}{10}
\providecommand{\url}[1]{#1}
\csname url@samestyle\endcsname
\providecommand{\newblock}{\relax}
\providecommand{\bibinfo}[2]{#2}
\providecommand{\BIBentrySTDinterwordspacing}{\spaceskip=0pt\relax}
\providecommand{\BIBentryALTinterwordstretchfactor}{4}
\providecommand{\BIBentryALTinterwordspacing}{\spaceskip=\fontdimen2\font plus
\BIBentryALTinterwordstretchfactor\fontdimen3\font minus \fontdimen4\font\relax}
\providecommand{\BIBforeignlanguage}[2]{{%
\expandafter\ifx\csname l@#1\endcsname\relax
\typeout{** WARNING: IEEEtran.bst: No hyphenation pattern has been}%
\typeout{** loaded for the language `#1'. Using the pattern for}%
\typeout{** the default language instead.}%
\else
\language=\csname l@#1\endcsname
\fi
#2}}
\providecommand{\BIBdecl}{\relax}
\BIBdecl

\bibitem{daw2022source}
A.~Daw, A.~Karpatne, K.~Yeo, and L.~Klein, ``Source identification and field reconstruction of advection-diffusion process from sparse sensor measurements.''\hskip 1em plus 0.5em minus 0.4em\relax Conference on Neural Information Processing Systems, 2022.

\bibitem{hon2010inverse}
Y.~Hon, M.~Li, and Y.~A. Melnikov, ``Inverse source identification by green's function,'' \emph{Engineering Analysis with Boundary Elements}, vol.~34, no.~4, pp. 352--358, 2010.

\bibitem{sung2023decision}
Y.~Sung, Z.~Chen, J.~Das, P.~Tokekar \emph{et~al.}, ``A survey of decision-theoretic approaches for robotic environmental monitoring,'' \emph{Foundations and Trends{\textregistered} in Robotics}, vol.~11, no.~4, pp. 225--315, 2023.

\bibitem{moghaddam2021inverse}
M.~B. Moghaddam, M.~Mazaheri, and J.~M.~V. Samani, ``Inverse modeling of contaminant transport for pollution source identification in surface and groundwaters: a review,'' \emph{Groundwater for Sustainable Development}, vol.~15, p. 100651, 2021.

\bibitem{yoerger2007autonomous}
D.~R. Yoerger, A.~M. Bradley, M.~Jakuba, C.~R. German, T.~Shank, and M.~Tivey, ``Autonomous and remotely operated vehicle technology for hydrothermal vent discovery, exploration, and sampling,'' \emph{Oceanography}, vol.~20, no.~1, pp. 152--161, 2007.

\bibitem{reddy2022olfactory}
G.~Reddy, V.~N. Murthy, and M.~Vergassola, ``Olfactory sensing and navigation in turbulent environments,'' \emph{Annual Review of Condensed Matter Physics}, vol.~13, no.~1, pp. 191--213, 2022.

\bibitem{vergassola2007infotaxis}
M.~Vergassola, E.~Villermaux, and B.~I. Shraiman, ``‘infotaxis’ as a strategy for searching without gradients,'' \emph{Nature}, vol. 445, no. 7126, pp. 406--409, 2007.

\bibitem{martin2010effectiveness}
E.~Martin~Moraud and D.~Martinez, ``Effectiveness and robustness of robot infotaxis for searching in dilute conditions,'' \emph{Frontiers in neurorobotics}, vol.~4, p. 1213, 2010.

\bibitem{hajieghrary2017information}
H.~Hajieghrary, D.~Mox, and M.~A. Hsieh, ``Information theoretic source seeking strategies for multiagent plume tracking in turbulent fields,'' \emph{Journal of Marine Science and Engineering}, vol.~5, no.~1, p.~3, 2017.

\bibitem{chen2025dlw}
B.~Chen, X.~Zhang, Y.~Ji, Y.~Zhao, and Z.~Zhu, ``Dlw-ci: A dynamic likelihood-weighted cooperative infotaxis approach for multi-drone cooperative multi-source search,'' \emph{Journal of Safety Science and Resilience}, 2025.

\bibitem{barbieri2011trajectories}
C.~Barbieri, S.~Cocco, and R.~Monasson, ``On the trajectories and performance of infotaxis, an information-based greedy search algorithm,'' \emph{EuroPhysics letters}, vol.~94, no.~2, p. 20005, 2011.

\bibitem{rodriguez2014limits}
J.~D. Rodr{\'\i}guez, D.~G{\'o}mez-Ullate, and C.~Mej{\'\i}a-Monasterio, ``Limits on the performance of infotaxis under inaccurate modelling of the environment,'' \emph{arXiv preprint arXiv:1408.1873}, 2014.

\bibitem{kinch2025structurepreservingdigitaltwinsconditional}
\BIBentryALTinterwordspacing
B.~Kinch, B.~Shaffer, E.~Armstrong, M.~Meehan, J.~Hewson, and N.~Trask, ``Structure-preserving digital twins via conditional neural whitney forms,'' 2025. [Online]. Available: \url{https://arxiv.org/abs/2508.06981}
\BIBentrySTDinterwordspacing

\bibitem{shaffer2025physicsinformedsensorcoveragestructure}
\BIBentryALTinterwordspacing
B.~D. Shaffer, B.~Kinch, J.~Klobusicky, M.~A. Hsieh, and N.~Trask, ``Physics-informed sensor coverage through structure preserving machine learning,'' 2025. [Online]. Available: \url{https://arxiv.org/abs/2509.10363}
\BIBentrySTDinterwordspacing

\bibitem{ogren2004cooperative}
P.~Ogren, E.~Fiorelli, and N.~E. Leonard, ``Cooperative control of mobile sensor networks: Adaptive gradient climbing in a distributed environment,'' \emph{IEEE Transactions on Automatic control}, vol.~49, no.~8, pp. 1292--1302, 2004.

\bibitem{Talwar24}
D.~Talwar, T.~Lu, D.~Sobti, and W.~Wu, ``Multi-robot source seeking and field reconstruction of spatial-temporal varying fields,'' in \emph{2024 9th International Conference on Automation, Control and Robotics Engineering (CACRE)}, 2024, pp. 101--105.

\bibitem{atanasov2012stochastic}
N.~Atanasov, J.~Le~Ny, N.~Michael, and G.~J. Pappas, ``Stochastic source seeking in complex environments,'' in \emph{2012 IEEE International Conference on Robotics and Automation}.\hskip 1em plus 0.5em minus 0.4em\relax IEEE, 2012, pp. 3013--3018.

\bibitem{al2021distributed}
S.~Al-Abri and F.~Zhang, ``A distributed active perception strategy for source seeking and level curve tracking,'' \emph{IEEE Transactions on Automatic Control}, vol.~67, no.~5, pp. 2459--2465, 2021.

\bibitem{zhangQin2023distributed}
T.~Zhang, V.~Qin, Y.~Tang, and N.~Li, ``Distributed information-based source seeking,'' \emph{IEEE Transactions on Robotics}, vol.~39, no.~6, pp. 4749--4767, 2023.

\bibitem{denniston2023fast}
C.~E. Denniston, O.~Peltzer, J.~Ott, S.~Moon, S.-K. Kim, G.~S. Sukhatme, M.~J. Kochenderfer, M.~Schwager, and A.-a. Agha-mohammadi, ``Fast and scalable signal inference for active robotic source seeking,'' in \emph{2023 IEEE International Conference on Robotics and Automation (ICRA)}.\hskip 1em plus 0.5em minus 0.4em\relax IEEE, 2023, pp. 7909--7915.

\bibitem{wang2019dynamic}
J.-W. Wang, Y.~Guo, M.~Fahad, and B.~Bingham, ``Dynamic plume tracking by cooperative robots,'' \emph{IEEE/ASME Transactions on Mechatronics}, vol.~24, no.~2, pp. 609--620, 2019.

\bibitem{singh2023emergent}
S.~H. Singh, F.~van Breugel, R.~P. Rao, and B.~W. Brunton, ``Emergent behaviour and neural dynamics in artificial agents tracking odour plumes,'' \emph{Nature machine intelligence}, vol.~5, no.~1, pp. 58--70, 2023.

\bibitem{zhang23}
Z.~Zhang, S.~T. Mayberry, W.~Wu, and F.~Zhang, ``Distributed cooperative kalman filter constrained by discretized poisson equation for mobile sensor networks,'' in \emph{2023 American Control Conference (ACC)}, 2023, pp. 1365--1370.

\bibitem{zhang2023distributed}
------, ``Distributed cooperative kalman filter constrained by advection--diffusion equation for mobile sensor networks,'' \emph{Frontiers in Robotics and AI}, vol.~10, p. 1175418, 2023.

\bibitem{mayberry2025soft}
S.~Mayberry, Z.~Zhang, W.~Wu, and F.~Zhang, ``Soft-constrained distributed cascaded cooperative kalman filter for mobile robots in unknown advection-diffusion field,'' \emph{IEEE Robotics and Automation Letters}, 2025.

\bibitem{guruswamy20}
S.~Guruswamy and W.~Wu, ``Cooperative level curve tracking in advection-diffusion fields,'' in \emph{2020 5th International Conference on Automation, Control and Robotics Engineering (CACRE)}, 2020, pp. 434--438.

\bibitem{khodayi2019model}
R.~Khodayi-mehr, W.~Aquino, and M.~M. Zavlanos, ``Model-based active source identification in complex environments,'' \emph{IEEE Transactions on Robotics}, vol.~35, no.~3, pp. 633--652, 2019.

\bibitem{khodayi2018physics}
R.~Khodayi-mehr and M.~M. Zavlanos, ``Physics-based learning for robotic environmental sensing,'' \emph{arXiv preprint arXiv:1812.03894}, 2018.

\bibitem{raissi2019physics}
M.~Raissi, P.~Perdikaris, and G.~E. Karniadakis, ``Physics-informed neural networks: A deep learning framework for solving forward and inverse problems involving nonlinear partial differential equations,'' \emph{Journal of Computational physics}, vol. 378, pp. 686--707, 2019.

\bibitem{lu2021learning}
L.~Lu, P.~Jin, G.~Pang, Z.~Zhang, and G.~E. Karniadakis, ``Learning nonlinear operators via deeponet based on the universal approximation theorem of operators,'' \emph{Nature machine intelligence}, vol.~3, no.~3, pp. 218--229, 2021.

\bibitem{li2020fourier}
Z.~Li, N.~Kovachki, K.~Azizzadenesheli, B.~Liu, K.~Bhattacharya, A.~Stuart, and A.~Anandkumar, ``Fourier neural operator for parametric partial differential equations,'' \emph{arXiv preprint arXiv:2010.08895}, 2020.

\bibitem{de2022deep}
M.~V. de~Hoop, M.~Lassas, and C.~A. Wong, ``Deep learning architectures for nonlinear operator functions and nonlinear inverse problems,'' \emph{Mathematical Statistics and Learning}, vol.~4, no.~1, pp. 1--86, 2022.

\bibitem{mishra2022estimates}
S.~Mishra and R.~Molinaro, ``Estimates on the generalization error of physics-informed neural networks for approximating a class of inverse problems for pdes,'' \emph{IMA Journal of Numerical Analysis}, vol.~42, no.~2, pp. 981--1022, 2022.

\bibitem{kamyab2022deep}
S.~Kamyab, Z.~Azimifar, R.~Sabzi, and P.~Fieguth, ``Deep learning methods for inverse problems,'' \emph{PeerJ Computer Science}, vol.~8, p. e951, 2022.

\bibitem{vesselinov2018contaminant}
V.~V. Vesselinov, B.~S. Alexandrov, and D.~O’Malley, ``Contaminant source identification using semi-supervised machine learning,'' \emph{Journal of contaminant hydrology}, vol. 212, pp. 134--142, 2018.

\bibitem{kontos2022machine}
Y.~N. Kontos, T.~Kassandros, K.~Perifanos, M.~Karampasis, K.~L. Katsifarakis, and K.~Karatzas, ``Machine learning for groundwater pollution source identification and monitoring network optimization,'' \emph{Neural Computing and Applications}, vol.~34, no.~22, pp. 19\,515--19\,545, 2022.

\bibitem{reiter2017machine}
A.~Reiter and M.~A.~L. Bell, ``A machine learning approach to identifying point source locations in photoacoustic data,'' in \emph{Photons Plus Ultrasound: Imaging and Sensing 2017}, vol. 10064.\hskip 1em plus 0.5em minus 0.4em\relax SPIE, 2017, pp. 504--509.

\bibitem{karniadakis2021physics}
G.~E. Karniadakis, I.~G. Kevrekidis, L.~Lu, P.~Perdikaris, S.~Wang, and L.~Yang, ``Physics-informed machine learning,'' \emph{Nature Reviews Physics}, vol.~3, no.~6, pp. 422--440, 2021.

\bibitem{wang2023physics}
S.~Wang, H.~Zhang, and X.~Jiang, ``Physics-informed neural network algorithm for solving forward and inverse problems of variable-order space-fractional advection--diffusion equations,'' \emph{Neurocomputing}, vol. 535, pp. 64--82, 2023.

\bibitem{trask2022enforcing}
N.~Trask, A.~Huang, and X.~Hu, ``Enforcing exact physics in scientific machine learning: a data-driven exterior calculus on graphs,'' \emph{Journal of Computational Physics}, vol. 456, p. 110969, 2022.

\bibitem{actor2024data}
J.~A. Actor, X.~Hu, A.~Huang, S.~A. Roberts, and N.~Trask, ``Data-driven whitney forms for structure-preserving control volume analysis,'' \emph{Journal of Computational Physics}, vol. 496, p. 112520, 2024.

\bibitem{jiang2024structure}
S.~Jiang, J.~Actor, S.~Roberts, and N.~Trask, ``A structure-preserving domain decomposition method for data-driven modeling,'' \emph{arXiv preprint arXiv:2406.05571}, 2024.

\bibitem{julian2014mutual}
B.~J. Julian, S.~Karaman, and D.~Rus, ``On mutual information-based control of range sensing robots for mapping applications,'' \emph{The International Journal of Robotics Research}, vol.~33, no.~10, pp. 1375--1392, 2014.

\bibitem{houlsby2011bayesian}
N.~Houlsby, F.~Husz{\'a}r, Z.~Ghahramani, and M.~Lengyel, ``Bayesian active learning for classification and preference learning,'' \emph{arXiv preprint arXiv:1112.5745}, 2011.

\bibitem{liu2021peril}
Y.~Liu, M.~Pagliardini, T.~Chavdarova, and S.~U. Stich, ``The peril of popular deep learning uncertainty estimation methods,'' \emph{arXiv preprint arXiv:2112.05000}, 2021.

\bibitem{gal2016dropout}
Y.~Gal and Z.~Ghahramani, ``Dropout as a bayesian approximation: Representing model uncertainty in deep learning,'' in \emph{international conference on machine learning}.\hskip 1em plus 0.5em minus 0.4em\relax PMLR, 2016, pp. 1050--1059.

\bibitem{arnold2018finite}
D.~N. Arnold, \emph{Finite element exterior calculus}.\hskip 1em plus 0.5em minus 0.4em\relax SIAM, 2018.

\bibitem{dimakis2010gossip}
A.~G. Dimakis, S.~Kar, J.~M. Moura, M.~G. Rabbat, and A.~Scaglione, ``Gossip algorithms for distributed signal processing,'' \emph{Proceedings of the IEEE}, vol.~98, no.~11, pp. 1847--1864, 2010.

\bibitem{bullo2009distributed}
F.~Bullo, J.~Cort{\'e}s, and S.~Martinez, \emph{Distributed control of robotic networks: a mathematical approach to motion coordination algorithms}.\hskip 1em plus 0.5em minus 0.4em\relax Princeton University Press, 2009.

\end{thebibliography}

\newpage
\section*{Appendix: Complex geometry}
\begin{figure}[htbp]
    \centering
    \includegraphics[width=0.22\textwidth]{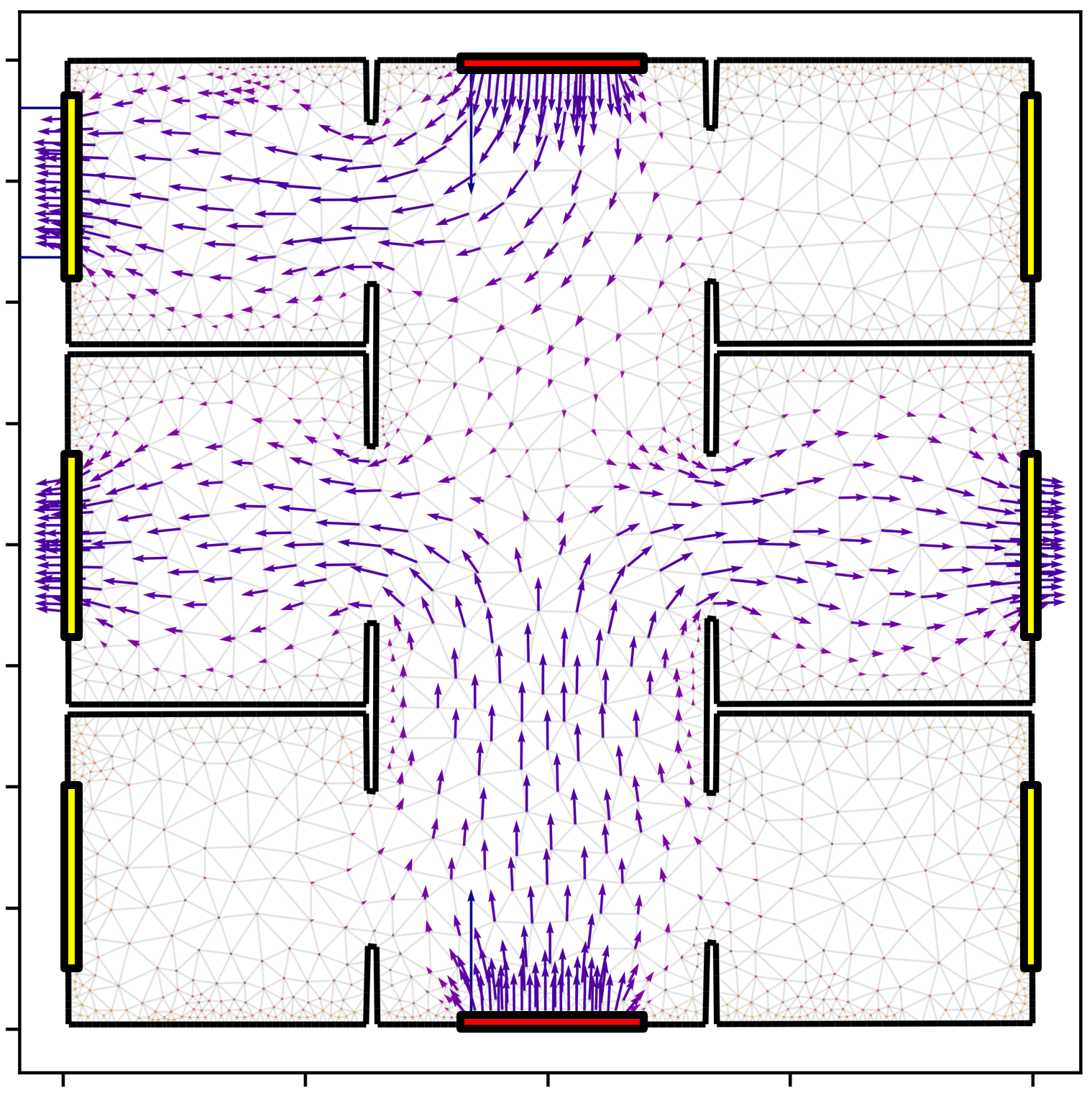}
    \caption{Our hallway-with-rooms geometry consists of two inlets (red) and six possible outlets (yellow). In the shown case three outlets are open.}
    \label{fig:complex_geom_demo}
\end{figure}

Figure~\ref{fig:complex_geom_demo} shows the domain and a sample flow field for the hallway geometry used in Section~\ref{sec:complex_geom}.

\end{document}